\pdfoutput=1

\documentclass[11pt]{article}

\usepackage{EACL2023}

\usepackage{times}
\usepackage{latexsym}

\usepackage[T1]{fontenc}

\usepackage[utf8]{inputenc}

\usepackage{microtype}

\usepackage{inconsolata}

\usepackage{xcolor,colortbl} 
\usepackage{amsmath} 
\usepackage{amssymb} 
\usepackage{wasysym} 
\usepackage{graphicx}

\definecolor{DarkBP}{RGB}{109, 157, 235}
\definecolor{MidBP}{RGB}{164, 194, 244}
\definecolor{LightBP}{RGB}{201, 218, 248}

\title{Text-Blueprint: An Interactive Platform for \\Plan-based
  Conditional Generation}

\author{
Fantine Huot, Joshua Maynez,
Shashi Narayan, \\ 
\textbf{Reinald Kim Amplayo,}  
\textbf{Kuzman Ganchev,} 
\textbf{Annie Louis,} \\ 
\textbf{Anders Sandholm,} 
\textbf{Dipanjan Das,} 
\textbf{Mirella Lapata} \\ Google Research \\ 
\texttt{\small fantinehuot@google.com, joshuahm@google.com, shashinarayan@google.com,} \\ \texttt{\small reinald@google.com, kuzman@google.com,  annielouis@google.com, } \\ 
\texttt{\small sandholm@google.com, dipanjand@google.com,
lapata@google.com}
}

\begin{document}
\maketitle

\begin{abstract}
While conditional generation models can now generate natural language
well enough to create fluent text, it is still difficult to control
the generation process, leading to irrelevant, repetitive, and
hallucinated content. Recent work shows that planning can be a useful
intermediate step to render conditional generation less opaque and
more grounded.  We present a web browser-based demonstration for
\mbox{query-focused} summarization that uses a sequence of question-answer
pairs, as a \emph{blueprint} plan for guiding text generation
(i.e.,~what to say and in what order). We illustrate how users may
interact with the generated text and associated plan visualizations,
e.g., by editing and modifying the blueprint in order to improve or control
the generated output.

A short video demonstrating our system is available at
\url{https://goo.gle/text-blueprint-demo}

\end{abstract}

\section{Introduction}

With the advent of encoder-decoder models
\cite{bahdanau2014neural,Sutskever-ea-2014}, Transformer-based
architectures \cite{NIPS2017_7181}, and large-scale pretraining
\cite{pmlr-v119-zhang20ae,lewis-etal-2020-bart},
deep learning models have achieved great performance on conditional
generation tasks such as summarization
\cite{rush-etal-2015-neural,nallapati-etal-2016-abstractive,see-etal-2017-get,liu-lapata-2019-text}
or task-oriented dialogue modeling
\cite{wen-etal-2018-sequence}. However, it remains challenging to
control the text generation, as these neural models tend to generate
hallucinated
\cite{song-etal-2018-structure,maynez-etal-2020-faithfulness,kryscinski-etal-2020-evaluating,gabriel-etal-2021-go}
or repetitive content
\cite{suzuki-nagata-2017-cutting,li-etal-2018-improving}, and struggle
to identify which information is most relevant to include in the
output text \cite{tan-etal-2017-abstractive}.

Recent work shows that planning can be a useful intermediate step to
address some of these challenges
\cite{DBLP:journals/corr/abs-1809-00582,moryossef-etal-2019-step,narayan-etal-2021-planning,narayan2022conditional}. In
this work, we present Text-Blueprint, a demonstration for showcasing
the approach described in \citet{narayan2022conditional}, that uses a
text plan, formulated as a sequence of question-answer pairs called
the \emph{blueprint}, to serve as an intermediate representation for
content selection and organization of the generated text. It draws
inspiration from the ``Questions Under Discussion'' theory of discourse
which posits that the structure of a text can be derived by identifying
the questions that are answered by each subsequent span of text
\cite{Carlson:1983,Ginzburg:1994,vanKuppevelt:1995,Larsson:2002,roberts:2012:information,Reister:2019}.

\begin{figure*}[th!]
\centering
\includegraphics[width=\textwidth,trim=100 250 100 50,clip]{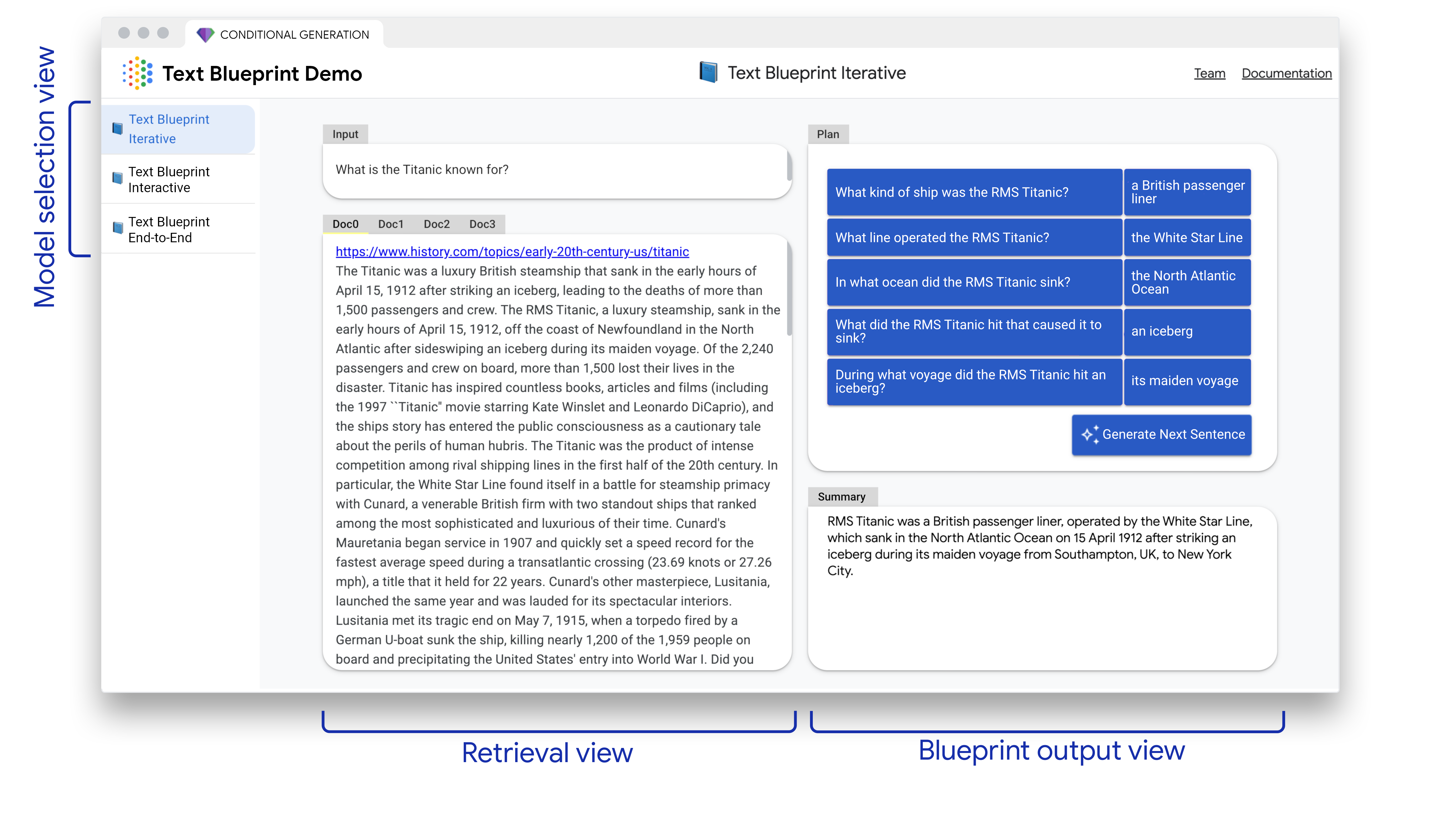}  
\caption{User interface of the web browser-based Text-Blueprint demonstration showcasing the iterative model.\label{ui}}
\end{figure*}

We implement this blueprint approach as an interactive web application
for query-focused summarization. An example snapshot of our interface
is shown in Figure \ref{ui}. As can be seen, for a given generated
summary, users can examine its corresponding blueprint, modify it to make it more
faithful or relevant, and control its length by changing the number of
question-answer pairs. Given  a query and relevant documents, there can be multiple
semantically-diverse summaries that meet the communicative goal of
synthesizing the most important points. Traditional generation systems do well at
\emph{single-best} summaries, while our interactive demonstration allows
users to explore \emph{different} summaries for a given input, while directly
observing the impact of changes to the plan on the generated text.  The
formulation of the blueprint plan as question-answer pairs makes it
intuitive and user-friendly (e.g., users can inspect and ask
questions without any instructions).

Our demonstration is an example of what can be achieved with
human-in-the-loop conditional generation \cite{cheng-etal-2022-mapping}. It
allows users to revise the output text (i.e., by editing the
blueprint) subject to their information needs. Additionally, it allows
researchers to analyze what constitutes a good blueprint for various
summarization tasks.

\section{Related Work}

There are  several libraries for broad NLP tasks, such as AllenNLP\footnote{\url{allennlp.org}} or GluonNLP\footnote{\url{gluon-nlp.mxnet.io}}. The Language Interpretability
Tool \cite{tenney2020language} is an interactive platform for examining model behavior, meant for rapid exploration and error analysis. A variety of toolkits have been developed recently that
support generation tasks. For instance, Texar \cite{hu-etal-2019-texar} is 
an open-source platform that
unifies the development of diverse yet closely-related applications, such as machine translation, summarization, and dialog.
TextBox \cite{li-etal-2021-textbox}  is a modular framework that offers
interfaces for various common functions in text generation models,
allowing researchers and practitioners to reproduce baseline models
and compare new models.  The Giant Language Model Test Room, also
known as GLTR \cite{gehrmann-etal-2019-gltr}, helps users
differentiate automatically-generated text from human-written text.

For conditional generation, many demonstrations are summarization systems. 
For instance, \citet{nyzam-bossard-2019-modular} present a modular tool for automatic summarization.
\citet{syed-etal-2021-summary} showcase a visualization tool for summaries obtained by different summarization methods. The SummVis
platform \cite{vig-etal-2021-summvis} serves a similar goal to the
demonstration presented in this paper. It enables users to visually
analyze the models, data, and evaluation metrics associated with
abtractive summarization, e.g., by highlighting hallucinated entities
in the generated text. While previous tools and frameworks are
versatile and modular, their focus is not on empowering users with
control over the generated text in an \emph{interactive} environment.  

In particular, studies on human-AI interaction for text summarization \cite{cheng-etal-2022-mapping,lai-etal-2022-exploration} show that users' overall  experience is better when they can control the generation process. Users preferred systems that allowed them to adjust and see the impact of their changes on the output directly, and the controllability improved their trust in the system when summarizing unfamiliar topics. 

Systems more geared toward interactive text generation include chatbots such as Meena \cite{adiwardana2020towards} or other specialized dialogue systems such as ParlAI \cite{miller2017parlai}. \citet{gehrmann2019visual} present an approach called collaborative semantic inference that exposes latent variables to the user for interactive generation. 
Still, these tasks differ from conditional generation using planning for which our demonstration is designed.

\section{Summarization Using Planning}

\begin{figure*}[th!]
\centering
\begin{tabular}{ c }
\includegraphics[width=\textwidth]{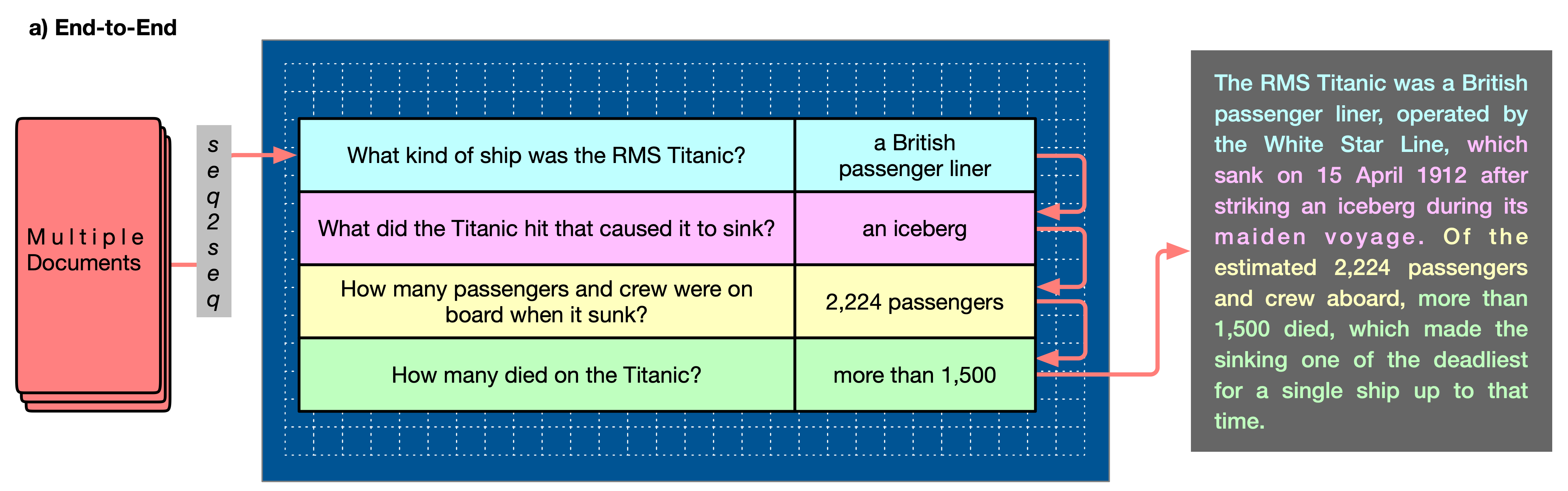} \\
\includegraphics[width=\textwidth]{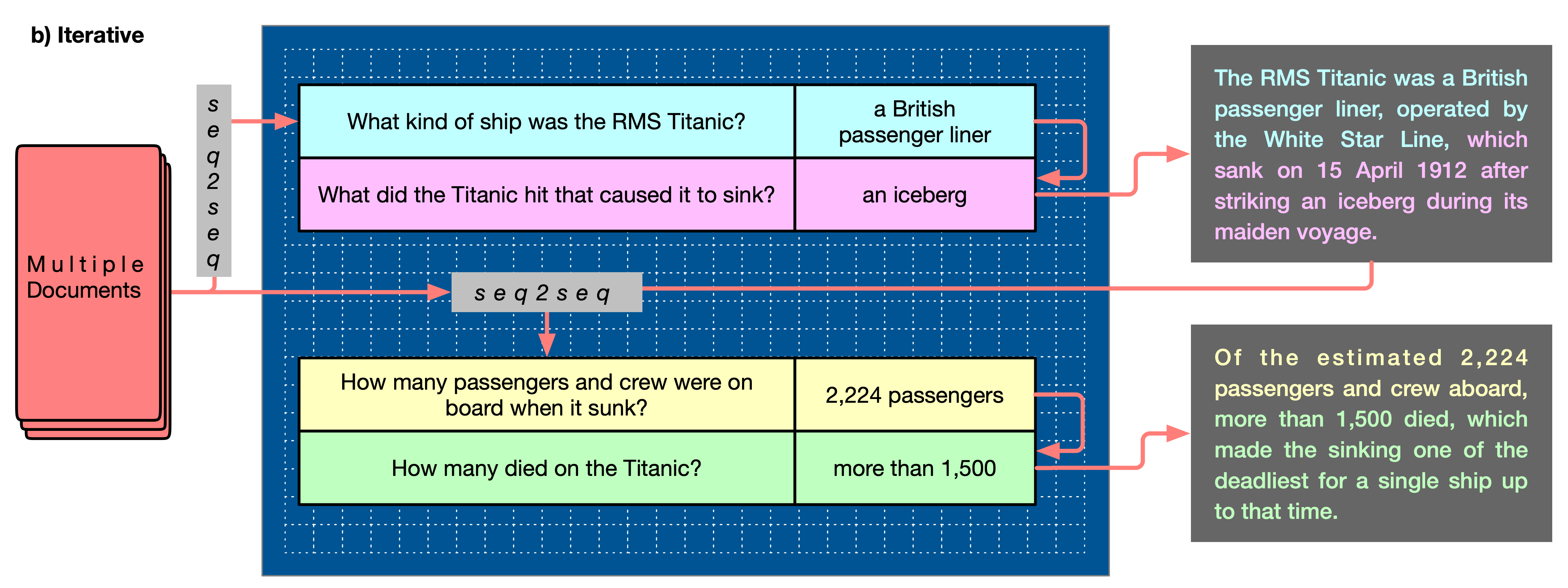}
\end{tabular}
\vspace{-0.2cm}
\caption{a) End-to-end and b) iterative Blueprint models. The end-to-end model generates the entire blueprint plan before generating the output text, while the iterative model plans and generates one proposition at a time, conditioning on the input and the sentences generated so far. Each portion of the output is color-coded with its corresponding question-answer pair. 
\label{fig:blueprint-variants} 
}
\end{figure*}

This demonstration showcases query-focused summarization using planning as described in \citet{narayan2022conditional}.\footnote{Code and checkpoints to be released at \url{github.com/google-research}} In their
approach, question-answer text plans, called blueprints, serve as
intermediate representations for content selection and structuring of
the generated text. We implement three Blueprint models in our
demonstration, which we briefly describe below; they are all
encoder-decoder variants instantiated from a Transformer
\cite{NIPS2017_7181} architecture.

Let~$d$ denote the input to our models, which is a user
query concatenated with a document or a set of documents relevant to
this query. From this input $d$, the model generates $b;s$, the
blueprint~$b$ and its corresponding summary~$s$. The blueprint itself
is a sequence of question-answer pairs $\{(q_1, a_1), (q_2, a_2),            
\ldots, (q_m, a_m)\}$. Existing datasets do not contain such
blueprints, they are typically designed as $(d, s)$
pairs. \citet{narayan2022conditional} describe a suite of data
enhancement methods for  obtaining blueprint annotations
(we refer the interested reader to their paper for details).

\paragraph{End-to-End Model}
The end-to-end Blueprint model uses an encoder-decoder model to encode the input documents~$d$ and generate~$b;s$, the concatenation of the blueprint and output text, \emph{in one go} (Figure~\ref{fig:blueprint-variants}a). The decoder first predicts the blueprint~$b$ autoregressively and continues to generate the output~$s$, conditioned on both~$b$ and~$d$. In particular, it predicts~$b$ as $q_1;a_1;\ldots;q_m;a_m$, a concatenated sequence of question-answer pairs. In contrast to systems that use prompted encoders, such as CTRLSum \cite{he2020ctrlsum}, the Blueprint models use decoder prompting. As a consequence, the blueprint plan is entirely generated by the model, without human intervention or relying on external systems. After the generation, users can inspect the question-answer pairs and corresponding summary. If desired, they can then select question-answer pairs to remove from the plan. The system is fed the updated blueprint~$b'$ which prompts the decoder to generate the corresponding output~$s’$.

\paragraph{Iterative Model}
It is generally challenging for encoder-decoder models to generate
long output sequences
\cite{ko-li-2020-assessing,tan-etal-2021-progressive}. The end-to-end
model ultimately suffers from this problem as it aims to generate
sequence $b;s$ instead of just $s$.
The iterative  Blueprint model mitigates this by adopting an incremental approach that interleaves planning with text generation rather than predicting a global plan before generating the output~$s$ (Figure~\ref{fig:blueprint-variants}b). 

If we denote the output~$s$ as consisting of $n$~sentences $\{s_1,s_2, \ldots, s_n\}$, then the corresponding blueprint~$b$ can be expressed as $\{b_1,b_2, \ldots, b_n\}$, where $b_i: \{(q_{1}^i, a_{1}^i), (q_{2}^i, a_{2}^i), \ldots, (q_{k}^i, a_{k}^i)\}$ consists of $k$ question-answer pairs for sentence~$s_i$. This model iteratively plans and generates one sentence at a time, conditioning on the input and the output sentences generated so far. In particular, it is trained such that the encoder first encodes the input~$d$, while the decoder takes the output generated so far $\{s_1, \ldots, s_i\}$ as a forced prompt and generates the blueprint $b_{i+1}$ for the following sentence $s_{i+1}$, followed by sentence~$s_{i+1}$ itself.

The iterative approach naturally addresses some of the issues the end-to-end model faces. In particular, it does not run into sequence length limitations as it predicts one sentence at a time.

\paragraph{Interactive Model}
The third model is an interactive Blueprint model that allows the user to modify the blueprint and directly change the generated output. It operates similarly to the end-to-end model previously described but in addition to letting the user select which elements of the plan to keep or remove, we design the system to allow the user to provide their own plan. 

We do not expect users to be able to provide answers to all the questions they come up with when creating their own plans. Therefore, we modify the original paradigm set by \citet{narayan2022conditional} and develop a new model specifically for the interactive mode that uses a question-only blueprint instead of a question-answer blueprint.  From the input documents $d$, we fine-tune this model to generate $b;s$, where $b$ is a concatenated sequence of questions $q_1;q_2;\ldots;q_m$. For this new model, we use the same blueprint training data as the iterative and end-to-end models, but only use the question annotations during fine-tuning, ignoring the answers. In the interactive mode, the user can edit the plan by typing in questions they come up with. This process creates an updated blueprint $b'$ which prompts the decoder to generate an updated summary $s’$.

\paragraph{Model Training}
The models made available in this demonstration are based on the LongT5 model  \cite{Guo:ea:2022}, an extension of T5 \cite{Raffel:ea:2019} designed to handle long inputs. 
Specifically, we fine-tune the XL 3B-parameter model\footnote{Using the checkpoints from \url{github.com/google-research/longt5}} with maximum input and output sequence lengths of 4,096 and 512 tokens, respectively, on the  AQuaMuSe dataset \cite{kulkarni2020aquamuse}. This is a query-focused multi-document summarization dataset which leverages the Google Natural Questions dataset \cite{kwiatkowski-etal-2019-natural}. The latter  contains real user queries from Google search logs paired with crowd-sourced answer spans from Wikipedia, and matched  with passages from web documents from Common Crawl. AQuaMuse uses the answer passages as summaries with the passages extracted from Common Crawl as the input documents. This dataset is query-focused, with long inputs and multi-sentence outputs, making it well-suited to a user-centric summarization system.

\section{System Description}

Our web browser-based demonstration is designed so that researchers and practitioners can inspect and interact with the different Blueprint models. We frame the summarization task around a user query, since in a real world scenario we would expect users to have a question or   intent in mind. The system retrieves documents relevant to the  query and displays their summary and its corresponding blueprint.
Figure~\ref{ui} provides a snapshot of the user interface (UI) and
 its components, namely the model selection, document retrieval, and
 Blueprint output views.

\paragraph{Model Selection View} Using the left-side menu, the user selects which of the models to use: end-to-end, iterative, or interactive Blueprint. The UI then adapts to the selected model. 

\paragraph{Retrieval View} In the search bar at the top of the middle panel, the user can enter an information-seeking query. For instance, in the example from Figure~\ref{ui}: ``What is the Titanic known for?''. 
The system automatically retrieves documents relevant to the query
and displays them underneath in different tabs, allowing the user to navigate between them and examine individual documents. The URL for each document is shown at the top. For longer documents, scrolling is also enabled. 

The document retrieval component is query-focused in a similar style
to the AQuaMuSe dataset \cite{kulkarni2020aquamuse}. It retrieves
candidate URLs and ranks relevant passages for a query using an
off-the-shelf retrieval system.\footnote{\url{github.com/google-research/t5x_retrieval}} It extracts a text
document from each of the best-ranking web pages, resulting in
multi-document input for the Blueprint models. Documents are formatted
similarly to AQuaMuse to closely match the data on which
the models were trained.

\paragraph{Blueprint Output View} The retrieved documents serve as inputs for the summarization. The outputs of the selected model are displayed on the right. The top-right box displays the blueprint $b$ and the bottom-right box shows the corresponding generated output $s$. The question-answer blueprint (or question-only, in the case of the interactive model) highlights what the model deemed important, such as, in the example from Figure~\ref{ui}, "What kind of ship is the Titanic?" or "What did the Titanic hit that caused it to sink?". We see that the generated output closely follows the blueprint.

In the end-to-end and interactive models, the user can click on elements of the blueprint to include or exclude them from the plan to re-generate the summary. Furthermore, when using the interactive model, an additional text box allows the user to input and edit a custom question plan.

\begin{figure}[t!]
\centering
\includegraphics[width=0.9\columnwidth]{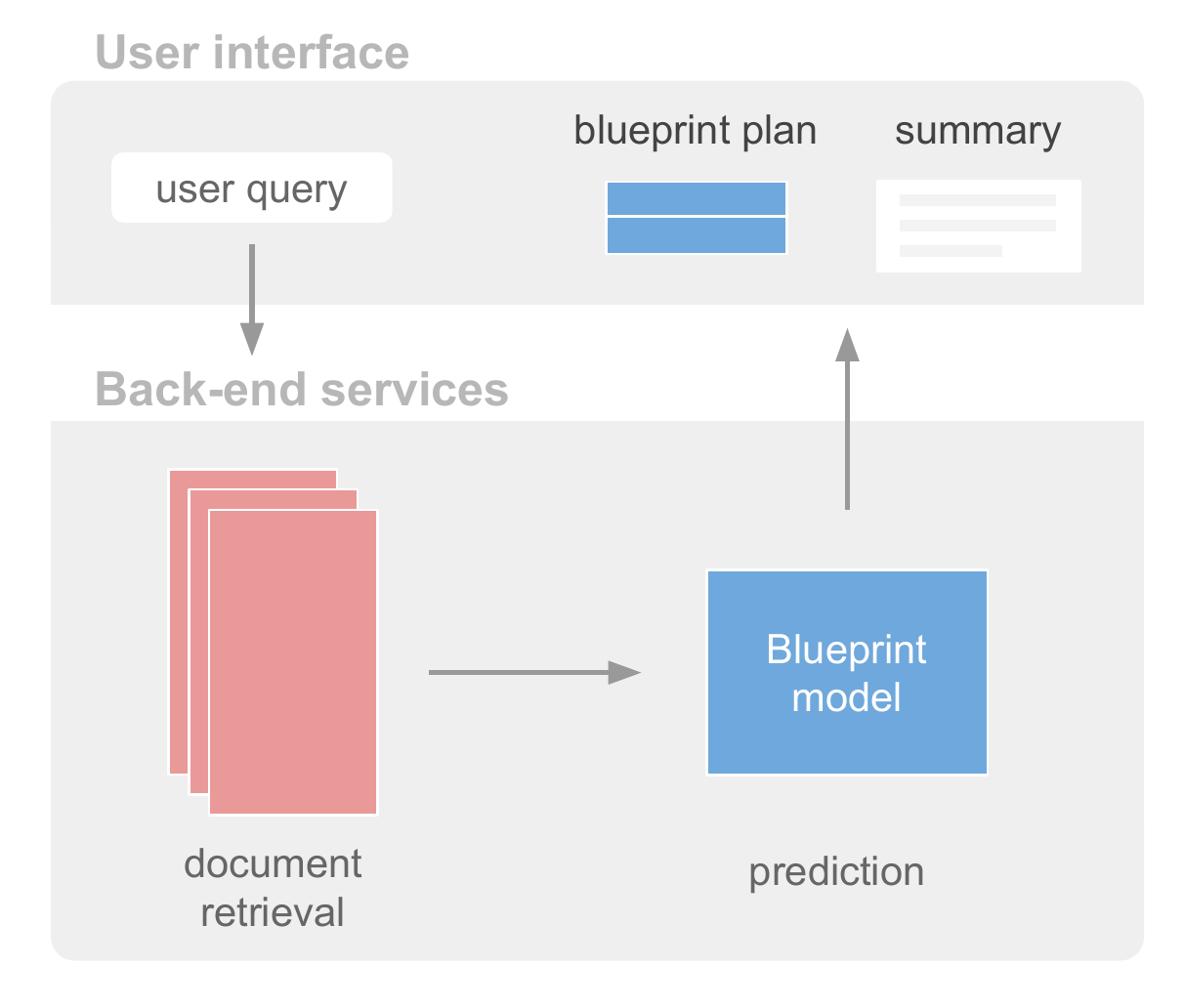} 
\caption{Schematic representation of the different components of the web browser-based demonstration. }
\label{fig:system}
\end{figure}

\paragraph{System Design}
Figure~\ref{fig:system} shows the different components of the web application. 
The web interface is made interactive with LitElement\footnote{See \url{lit.dev} for details.} components and is implemented in HTML and TypeScript. The back-end services are implemented in Python and~C++. Requests for document retrieval and blueprint model inference are sent to  back-end services to be processed asynchronously. Outputs are then sent back to the front-end web interface.

\section{Use Cases}

In the following we explore some of the possibilities of human-in-the-loop summarization and  illustrate different use cases for our demonstration.

\begin{figure*}[t]
\centering
\includegraphics[width=\textwidth]{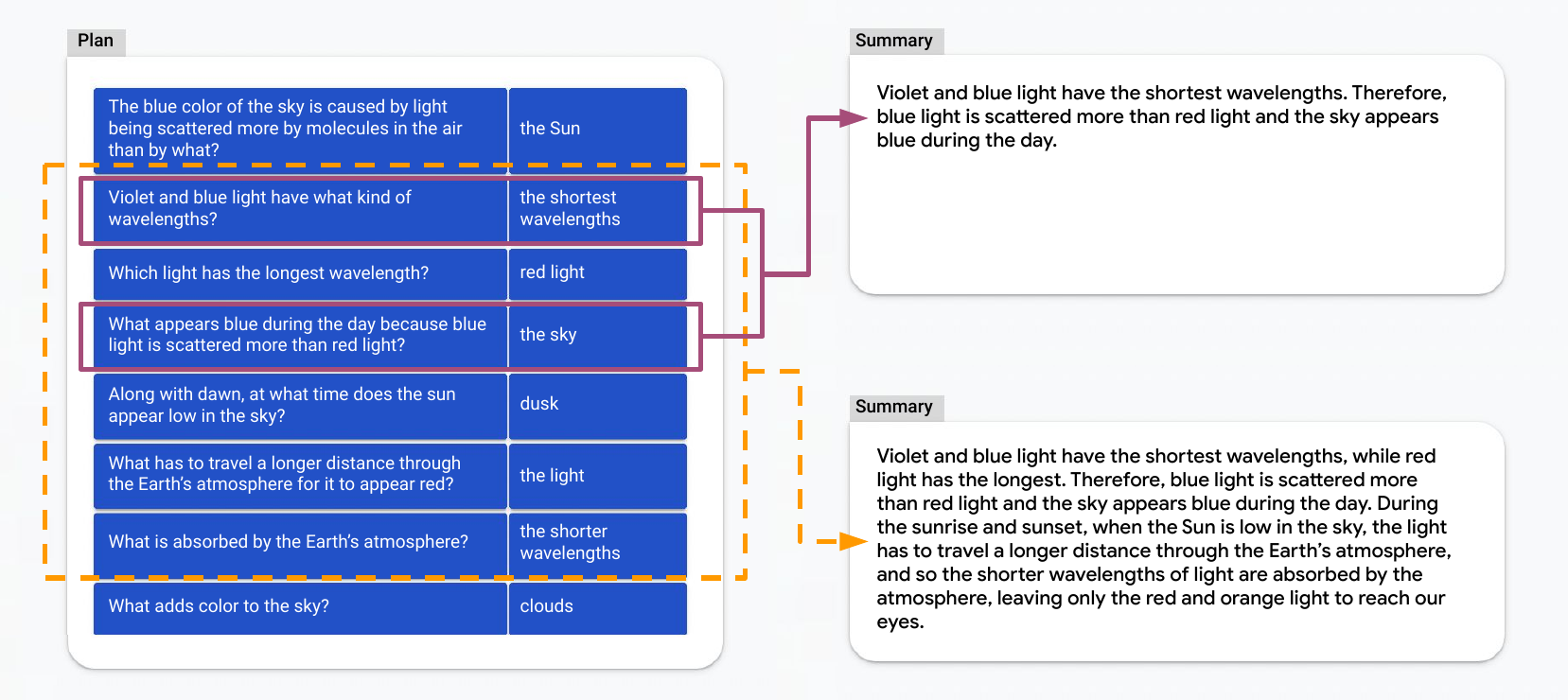}
\caption{Example snapshot of the results obtained with the end-to-end Blueprint model for the user query "Why is the sky blue?". Depending on which question-answer pairs the user selects, different summaries can be generated. 
}
\label{fig:blue_sky}
\end{figure*}

\paragraph{Informative Blueprints}
While the inner workings of deep learning models might be opaque to a human user, the formulation of the blueprint as a sequence of questions makes the control of the system's output user-friendly. Users do not have to be machine learning experts to interact with the system through questions and answers. Moreover, the ability to change the blueprint and observe the result on the summary, provides the user with immediate feedback.

The planning step also brings some insight into the often black-box nature of conditional generation. This property is especially valuable when the user summarizes complex or difficult information, since it breaks down the generation process into a sequence of questions. The blueprint plan offers context for the information in the generated output, which has been shown to be a desirable property in human-AI interaction for text summarization \cite{cheng-etal-2022-mapping}. Plan agnostic models do not provide details as to why certain pieces of information were included. In contrast, as seen in Figure~\ref{fig:blue_sky}, the blueprint plan anchors conditional generation, providing the user with a question-answer explanation for each proposition. 

\paragraph{Improved Faithfulness}  \citet{narayan-etal-2021-planning} evaluate the Blueprint models across several datasets and show improvements in faithfulness over models that do not use planning. Moreover, they also evaluate the impact of automated blueprint edits on the output summary. For each generated blueprint, they automatically remove  question-answer pairs for which the answer is not contained in the input, thus eliminating questions that cannot be answered based on the input documents. They then prompt the decoder with the modified blueprint to generate the summary, following a similar setting as in our system demonstration. Their results confirm that this automatic filtering of the generated blueprint further improves faithfulness. Their experiment further underscores  the importance of letting users interactively modify the plan, since we expect manual editing of the blueprint to have greater potential than automated filtering alone. In addition to unanswerable questions, users can remove questions with incorrect answers and irrelevant questions. A quantitative evaluation of the full scope of various human edits (e.g.,~remove an element of the blueprint, reorder the blueprint, add human-written questions) is left for future work.

\paragraph{Controllable Blueprints} 
In the example in Figure \ref{fig:blue_sky}, we examine the blueprint results obtained with the end-to-end model for the user query ``Why is the sky blue?''. The first question-answer pair of the  blueprint is actually incorrect, but likely would not have been caught by simple heuristics since it seems fluent and its answer is present in the input documents. The user selects the subset of question-answers that are deemed most relevant, leading to  higher-quality output than would have been generated without the blueprint control step. In particular, we see that the output does not contain the inaccuracies from the first question-answer pair. 

This example also shows how the user can control the length of the generated summary by including more or less question-answer pairs in the blueprint. For instance, the user can restrict the summary to contain only the explanation for blue skies as shown at the top in Figure~\ref{fig:blue_sky}, or decide to include  information about orange skies at sunset, as shown at the bottom. For a given query and source documents, the system can lead to diverse summaries by selecting different blueprints. Moreover, while it would be difficult for a user to come up with a plan from scratch if they are unfamiliar with the topic of their query, the provided blueprint can serve as a starting point from which the user can select what they would like to keep. As we discuss next, the user could also  elaborate on their initial query by adding  questions in the blueprint. 

\begin{figure*}[t!]
\centering
\includegraphics[width=\textwidth]{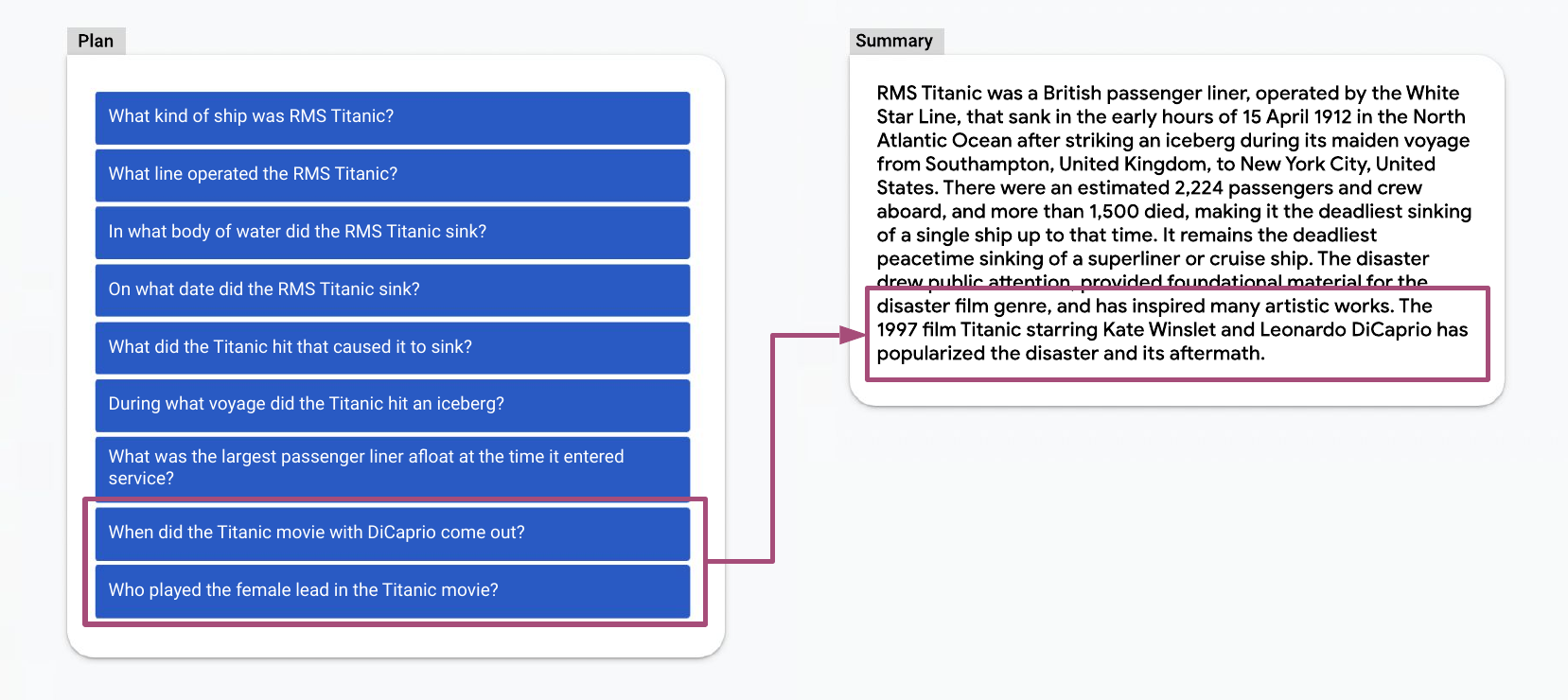}
\caption{Snapshot of the results obtained with the interactive Blueprint model for the  query ``What is the Titanic known for?''. Questions highlighted in red were manually added by the user, leading to a different output.
}
\label{fig:titanic}
\end{figure*}

\paragraph{Personalized Generation} Going beyond selecting and removing questions, in Figure~\ref{fig:titanic}, we illustrate results obtained with the interactive model and a user-provided blueprint. The user can edit the blueprint with their own follow-on questions, leading to an updated summary with information  it did not contain originally. When summarizing unfamiliar topics, it might be difficult for the user to come up with many new questions, and such cases might be better served by the end-to-end model. In Appendix~\ref{sec:appendix}, we provide additional examples of manually-edited blueprints and their corresponding summary. We observe that editing the plan allows the user to guide the generation to include certain elements in the output summary. 

\section{Conclusions}
This demonstration showcases a novel approach to query-focused summarization that uses a blueprint to plan the generated text. By implementing it within an  interactive framework, we transform it into an example of human-in-the-loop conditional generation. Our demonstration  is designed in a query-focused summarization setting; it retrieves multiple documents for a given query and uses them
as input for the summarization. The system offers three different model variations, namely an end-to-end,  iterative, and  interactive Blueprint approach. The interactive model, in particular, allows users to examine and edit the blueprint plans, offering a more personalized experience. Since the blueprint is formulated as a sequence of questions, it provides a natural way for the user to interact with the generated output,  e.g., by  selecting  relevant question-answer pairs, which in turn helps reduce inaccuracies and  hallucinations.

We hope this demonstration will spur further exploration into controllable and interpretable conditional generation systems and how human interaction can be an integral component in generating personalized outputs.
We further expect interactive tools like the one presented here to assist
in summary creation and editing, e.g., for data augmentation in low-resource settings or for more robust system evaluation by generating multiple outputs for a given document.

\section{Ethics statement} 
An ethical consideration with generative language models is the problem of misinformation. While the work we present here makes a step towards improving the faithfulness and factual consistency of text generation systems, it is important to note that current systems are still far from being perfect in this respect, and thus should be used with caution. While we did not observe harmful speech with typical queries, such a system can still be abused and additional controls and filters on both the queries and the system's output could help mitigate this.

\section*{Acknowledgements}
We thank Sebastian Gehrmann, Ankur Parikh, and William Cohen for their feedback on earlier versions of this work. 

\bibliography{anthology,custom,qud}
\bibliographystyle{acl_natbib}

\appendix

\section{Appendix}
Tables \ref{table:examples1} and \ref{table:examples2} on the following pages show  examples of manually-edited plans and their corresponding summaries.
\label{sec:appendix}

\begin{table*}[t]
\resizebox{\linewidth}{!}{
\setlength{\tabcolsep}{7pt}%
  \begin{tabular}{cll}

\multicolumn{3}{l}{Why did France give the US the Statue of Liberty?} \\ \hline
\\
\rowcolor{LightBP} \cellcolor{white} \rotatebox[origin=c]{90}{\textsc{Machine-Generated Plan}} & \parbox{12cm}{Q$_{1}$: Who proposed that a statue be built as a gift from France to the United States to commemorate the friendship between France and the United States? \\ Q$_{2}$: In what year was the Statue of Liberty designed? \\ Q$_{3}$: Who designed the Statue of Liberty? \\ Q$_{4}$: Along with freedom and democracy, what did Laboulaye want the Statue of Liberty to represent? \\ Q$_{5}$: To whom was the Statue of Liberty presented? \\ Q$_{6}$: In what city was the Statue of Liberty presented to the U.S. ambassador? \\ Q$_{7}$: Gustave Eiffel helped build what for the Statue of Liberty?} & \cellcolor{white} \parbox{12cm}{In 1865, French political thinker and abolitionist Édouard de Laboulaye proposed that a statue be built as a gift from France to the United States to commemorate the friendship between France and the United States and to commemorate the 100th anniversary of U.S. independence. Designed by sculptor Auguste Bartholdi, the statue was meant to symbolize the American welcome of immigrants and refugee seekers. It was presented to the U.S. ambassador Levi Morton in Paris on July 4, 1881, by French Marshal Nicolas Luckner as a commemoration of the friendship between France and the United States. Gustave Eiffel helped build the pedestal for the Statue of Liberty.} \\
\\
\rowcolor{LightBP} \cellcolor{white} \rotatebox[origin=c]{90}{ \textsc{Manually-Edited Plan}} & \parbox{12cm}{Q$_{1}$: Who proposed that a statue be built as a gift from France to the United States to commemorate the friendship between France and the United States? \\ Q$_{2}$: In what year was the Statue of Liberty designed? \\ Q$_{3}$: Who designed the Statue of Liberty? \\ Q$_{4}$: Along with freedom and democracy, what did Laboulaye want the Statue of Liberty to represent? \\ Q$_{5}$: To whom was the Statue of Liberty presented? \\ Q$_{6}$: In what city was the Statue of Liberty presented to the U.S. ambassador? \\ \textbf{Q$_{7}$: How was the Statue of Liberty transported to New York City?}} & \cellcolor{white} \parbox{12cm}{In 1865, French political thinker and abolitionist Édouard de Laboulaye proposed that a statue be built as a gift from France to the United States to commemorate the friendship between France and the United States and to commemorate the 100th anniversary of the United States' independence. Designed by sculptor Auguste Bartholdi, the statue was meant to symbolize the American welcome of immigrants and refugee seekers. It was presented to the U.S. ambassador Levi Morton in Paris on July 4, 1881, \textbf{and later transported to New York City where it was reconstructed.}} \\
\\
\\
\multicolumn{3}{l}{Who plays Queen Lannister in Game of Thrones?} \\ \hline
\\
\rowcolor{LightBP} \cellcolor{white} \rotatebox[origin=c]{90}{ \textsc{Machine-Generated Plan}} & \parbox{12cm}{Q$_{1}$: Who is the actress who played Cersei Lannister? \\ Q$_{2}$: What is Cersei Lannister's role in Game of Thrones? \\ Q$_{3}$: What is the name of the book series in which Cersei Lannister appears? \\ Q$_{4}$: Who wrote 'A Song of Ice and Fire'? \\ Q$_{5}$: What is Game of Thrones based on? \\ Q$_{6}$: When did Game of Thrones first air? \\ Q$_{7}$: What is one of the wealthiest and most powerful families in Westeros? \\ Q$_{8}$: In what year did Cersei Lannister first appear in the books?} & \cellcolor{white} \parbox{12cm}{Cersei Lannister is a fictional character in the A Song of Ice and Fire series of epic fantasy novels by American author George R.R. Martin, and its television adaptation Game of Thrones, where she is portrayed by Bermudian-English actress Lena Headey. Introduced in 1996's A Game of Thrones, Cersei is a member of House Lannister, one of the wealthiest and most powerful families on the continent of Westeros. She subsequently appears in A Clash of Kings (1998) and A Storm of Swords (2000), and becomes a prominent point of view character beginning with A Feast for Crows (2005).} \\
\\
\rowcolor{LightBP} \cellcolor{white} \rotatebox[origin=c]{90}{ \textsc{Manually-Edited Plan}} & \parbox{12cm}{Q$_{1}$: Who is the actress who played Cersei Lannister?\\ Q$_{2}$: What is Cersei Lannister's role in Game of Thrones? \\ \textbf{Q$_{3}$: Has the actress received any awards? \\ Q$_{4}$: Is the actress famous?}} & \cellcolor{white} \parbox{12cm}{Cersei Lannister is a fictional character in the A Song of Ice and Fire series of epic fantasy novels by American author George R.R. Martin, and its television adaptation Game of Thrones, where she is portrayed by Bermudian-English actress Lena Headey. \textbf{Headey has received widespread critical acclaim for her portrayal of the character, making her one of the most popular and longest-running actors on television.}} \\
\\
\\
\multicolumn{3}{l}{Who said my enemy's enemy is my friend?} \\ \hline
\\
\rowcolor{LightBP} \cellcolor{white} \rotatebox[origin=c]{90}{ \textsc{Machine-Generated Plan}} & \parbox{12cm}{Q$_{1}$: What did the Latin phrase "Amicus meus, inimicus inimici mei" mean? \\ Q$_{2}$: During what time period did the Latin phrase "Amicus meus, inimicus inimici mei" become common? \\ Q$_{3}$: When did the English version of the phrase "the enemy of my enemy is my friend" first appear? \\ Q$_{4}$: Who was the first person to use the modern English version of the phrase "the enemy of my enemy is my friend"? \\ Q$_{5}$: When did the English version of the phrase "the enemy of my enemy is my friend" first appear?} & \cellcolor{white} \parbox{12cm}{The exact meaning of the modern phrase was first expressed in the Latin phrase "Amicus meus, inimicus inimici mei" ("my friend, the enemy of my enemy"), which had become common throughout Europe by the early 1700s, while the first recorded use of the current English version came in 1884. The first recorded instance for this phrasing comes from Gabriel Manigault, who in his 1884 Political Creed described the sense that "the enemy of my enemy is my friend" as a "natural feeling".} \\
\\
    \rowcolor{LightBP} \cellcolor{white} \rotatebox[origin=c]{90}{ \textsc{Manually-Edited Plan}} &\parbox{12cm}{\textbf{Q$_{1}$: Who was the first person to use the modern English version of the phrase "the enemy of my enemy is my friend"? \\ Q$_{2}$: What did Gabriel Manigault describe as a natural feeling?}} & \cellcolor{white} \parbox{12cm}{\textbf{The first recorded instance for this phrasing comes from Gabriel Manigault, who in his 1884 Political Creed described the sense that "the enemy of my enemy is my friend" as a "natural feeling".}} \\
\\
\end{tabular}}
\caption{Examples of machine-generated and manually-edited plans and their corresponding summaries. We highlight in bold changes made by the user and the resulting changes to the summary. \label{table:examples1}
}
\end{table*}

\begin{table*}[t]
\resizebox{\linewidth}{!}{
\setlength{\tabcolsep}{7pt}%
  \begin{tabular}{cll}

\multicolumn{3}{l}{Is it illegal to have a bald eagle?} \\ \hline
\\
\rowcolor{LightBP} \cellcolor{white} \rotatebox[origin=c]{90}{ \textsc{Machine-Generated Plan}} & \parbox{12cm}{Q$_{1}$: What is the section number of the Bald and Golden Eagle Protection Act in the United States Code? \\ Q$_{2}$: What is the Bald and Golden Eagle Protection Act? \\ Q$_{3}$: What does the Bald and Golden Eagle Protection Act prohibit the taking of? \\ Q$_{4}$: Along with golden eagles, what bald eagle is protected by the Bald and Golden Eagle Protection Act? \\ Q$_{5}$: The Bald and Golden Eagle Protection Act prohibits the taking of bald eagles, their parts, nests, and what else? \\ Q$_{6}$: Who is required to obtain a permit from the Secretary of the Interior to take bald eagles? \\ Q$_{7}$: What can result in a fine of up to \$100,000? \\ Q$_{8}$: How much is the fine for a first offense of taking a bald eagle?
} & \cellcolor{white} \parbox{12cm}{The Bald and Golden Eagle Protection Act (16 U.S.C. 668-668d), enacted in 1940, and amended several times since, prohibits anyone, without a permit issued by the Secretary of the Interior, from "taking" bald or golden eagles, including their parts (including feathers), nests, or eggs. The Act provides criminal penalties for persons who take, possess, sell, purchase, barter, offer to sell, purchase or barter, transport, export or import, at any time or any manner, any bald eagle (or any golden eagle), alive or dead, or any part (including feathers), nest, or egg thereof. A violation of the Act can result in a fine of up to \$100,000 (\$200,000 for organizations), imprisonment for one year, or both, for a first offense.} \\
\\
\rowcolor{LightBP} \cellcolor{white} \rotatebox[origin=c]{90}{ \textsc{Manually-Edited Plan}} & \parbox{12cm}{Q$_{1}$: What is the Bald and Golden Eagle Protection Act? \\ \textbf{Q$_{2}$: Can Native Americans ask for an eagle permit?}} & \cellcolor{white} \parbox{12cm}{The Bald and Golden Eagle Protection Act (16 U.S.C. 668-668d), enacted in 1940, and amended several times since, prohibits anyone, without a permit issued by the Secretary of the Interior, from "taking" bald or golden eagles, including their parts (including feathers), nests, or eggs. \textbf{Native Americans may ask for an eagle permit.}} \\
\\
\\
\multicolumn{3}{l}{Is software engineer a good job?} \\ \hline
\\
\rowcolor{LightBP} \cellcolor{white} \rotatebox[origin=c]{90}{ \textsc{Machine-Generated Plan}} & \parbox{12cm}{Q$_{1}$: What is projected to grow 22\% from 2020 to 2030? \\ Q$_{2}$: What is the average salary for a software engineer? \\ Q$_{3}$: What is the average salary for a software engineer? \\ Q$_{4}$: Along with management, in what area do software engineers earn more than most other workers?} & \cellcolor{white} \parbox{12cm}{Employment of software developers is projected to grow 22\% from 2020 to 2030, which is much higher than the national average for other occupations. The average salary for a software engineer is \$99,400 with an average yearly growth rate of 7\%. In addition, software engineers earn more than most other workers in the more traditional business aspects such as management and sales.} \\
\\
\rowcolor{LightBP} \cellcolor{white} \rotatebox[origin=c]{90}{ \textsc{Manually-Edited Plan}} & \parbox{12cm}{Q$_{1}$: What is the average salary for a software engineer? \\ \textbf{Q$_{2}$: What degree should you get to become a software engineer? }} & \cellcolor{white} \parbox{12cm}{The average salary for a software engineer is \$99,400 according to the BLS. \textbf{Having a bachelor's degree in computer science or software engineering is recommended, though a master's degree may be more beneficial.}} \\
\\
\end{tabular}}
\caption{Examples of machine-generated and manually-edited plans and their corresponding summaries (Continued). We highlight in bold changes made by the user and the resulting changes to the summary. \label{table:examples2}
}
\end{table*}
\end{document}